\documentclass{INTERSPEECH2023}
\usepackage{tablefootnote}
\usepackage{tabularx}
\usepackage[table]{xcolor}

\usepackage{color, colortbl}
\definecolor{Gray}{RGB}{192,192,192}
\definecolor{Question}{RGB}{230,179,195}
\definecolor{Rationale}{RGB}{229,240,219}

\definecolor{Question Description}{RGB}{234,180,138}
\definecolor{History}{RGB}{255,217,102}

\interspeechcameraready

\title{Prompt Guided Copy Mechanism for Conversational Question Answering}
\name{Yong Zhang, Zhitao Li, Jianzong Wang*, Yiming Gao, Ning Cheng,  Fengying Yu, Jing Xiao\thanks{*Corresponding author: Jianzong Wang, jzwang@188.com.}}
\address{
  Ping An Technology (Shenzhen) Co., Ltd., China}

\email{  }

\begin{document}

\maketitle
 
\begin{abstract}

Conversational Question Answering (CQA) is a challenging task that aims to generate natural answers for conversational flow questions. In this paper, we propose a pluggable approach for extractive methods that introduces a novel prompt-guided copy mechanism to improve the fluency and appropriateness of the extracted answers. Our approach uses prompts to link questions to answers and employs attention to guide the copy mechanism to verify the naturalness of extracted answers, making necessary edits to ensure that the answers are fluent and appropriate. The three prompts, including a question-rationale relationship prompt, a question description prompt, and a conversation history prompt, enhance the copy mechanism's performance. Our experiments demonstrate that this approach effectively promotes the generation of natural answers and achieves good results in the CoQA challenge. 
\end{abstract}


\section{Introduction}

Conversational Question Answering (CQA) aims to answer conversation flow questions given an understanding of the text. Unlike traditional single-turn question answering, CQA questions are conversational in nature. They are based on the conversation history, which may require history modeling to handle coreference and pragmatic reasoning. In addition, CQA answers are coherent and must consider the previous questions and answers in the conversation. This coherence means that pure rationale may need to be edited for the natural flow of the conversation.

Extractive methods focus on identifying the relevant rationale from the passage to answer the question, based on the connection between the question and the conversation history. These methods employ an extraction-based approach to generate answers, by predicting the start and end position of the rationale in the passage.  HAE\cite{Hae} considers the construction of historical answer markers in the embeddings. FlowQA\cite{FlowQA} mimics the memory mechanism of humans by processing questions and information sequentially as the conversation progresses. GHR\cite{GHR} introduces a dialogue-based attention mechanism to capture interactions between different conversation turns, and Excord\cite{Excord} rewrites the questions based on the history to complement meaning before treating them as single-turn QA. However, these methods often ignore the relationship between the question and the passage rationale. 

On the other hand, naturalness-based methods focus on generating fluent and coherent answers, using the sequence-to-sequence structure. Answer-BART\cite{answer_bart} uses an end-to-end model to process the question and passage, generating potential evidence and a natural answer. REAG\cite{reag} incorporates the evidence extraction task into the transformer model's encoder to improve the natural answer's confidence. Unlike the above methods, S-net\cite{s_net} fuses the extraction and generation that it first uses the extraction model to collect the passage's most-important sub-text and then synthesize them into the final answer by the generative model. 

\begin{figure*}[!tp]

\begin{center}
\includegraphics[width=0.83\textwidth]{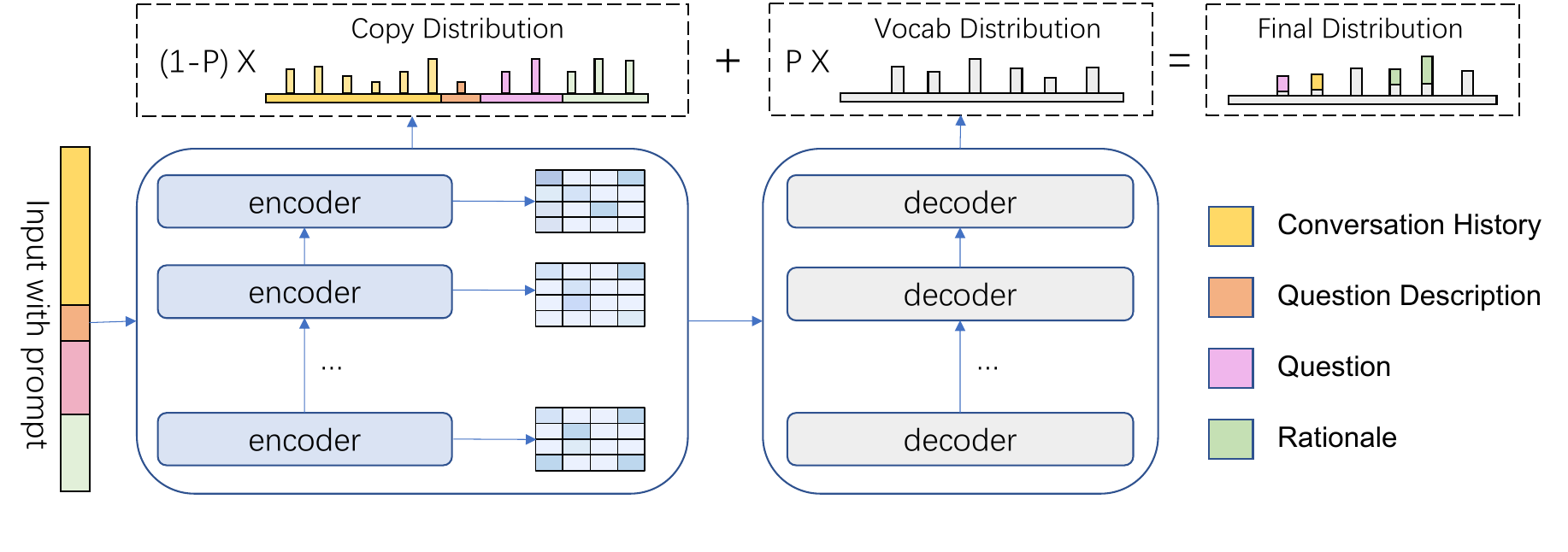}
\end{center}    
\vspace{-8.5mm}

\caption{The architecture of our proposed model}
\label{run}
\vspace{-6mm}

\end{figure*}

Although sequence-to-sequence methods have achieved significant success in generating natural and fluent answers in conversational question answering, they still suffer from the risk of producing hallucinated answers that are not related to the input. In contrast, extractive methods can directly capture answer spans from the associated text, which reduces the randomness of generative models and provides a strong connection to the input. However, extractive methods often do not take the conversation history and question type into account.

We propose a pluggable prompt-guided copy mechanism to improve the naturalness of extractive answers in conversational question answering. The mechanism verifies and potentially rewrites the extracted rationale span to meet the two principles of naturalness. Firstly, the answer should be fluent, given that the rationale text may be partial, redundant, and not always consistent with the question. Secondly, the answer should be appropriate for the question type, as the natural rationale text may need to be rephrased to suit the question type.

To improve the copy mechanism, we propose the use of three prompts to link conversational questions with rationale texts. These prompts include a prompt that establishes the relationship between the question and its rationale, a prompt that provides additional information about the question, and a prompt that draws on examples from past conversations. While prompts are commonly used in single-turn question answering\cite{liu2021gpt,reina,proqa}, our approach applies them to multi-turn CQA tasks for more effective fine-tuning.

During fine-tuning, explicit text prompt clues are transformed into hidden embeddings and fully interacted under the self-attention mechanism to activate crucial words. These embeddings are then converted into guidance information for the copy mechanism to make better decisions in generating the answer. To leverage the information at different granularities constructed in the prompts, we propose letting the decoder pay attention to all representations generated by the encoders, providing a multi-view perspective.

Our paper makes the following contributions:
\begin{itemize}
    \item
 We propose a pluggable approach for extractive methods that uses a prompt-guided copy mechanism to improve the fluency and appropriateness of extracted answers.

    \item

We propose the use of three prompts, including a question-rationale relationship prompt, a question description prompt, and a conversation history prompt, to enhance the effectiveness of the copy mechanism.
    \item
    We introduce a multi-view perspective attention mechanism that digests prompts to guide the copy mechanism and improve its performance.

    \item
    We demonstrate the effectiveness of the proposed approach in promoting the generation of natural answers and achieving good results in the CoQA\cite{reddy2019coqa} challenge.
\end{itemize}

\vspace{-2mm}

\section{Methodology}
\label{sec:method}
\vspace{-1mm}


This section describes our prompt-guided copy mechanism method to enhance the naturalness of answers in conversational question answering. Figure 1 outlines the method's architecture, highlighting the interplay between the question and the rationale span in final answer generation.

The method transforms raw inputs into enriched text through tailored prompts, which are converted into embeddings by the encoder. This enriched embedding guides the decoder's copy mechanism to generate the final natural answer.



Formally, an ``answer" is a consistent response in dialogues that follows the right reference, tense, and grammar, and matches the question format. The ``rationale" is a small text piece from the reference article, which may be semantically incomplete and is not bound by the questions in the dialogue. We consider extracted answers as rationales. Given a question $q_t$ and rationale $r_t$ with history dialogue $H_t=[(q_{1}, a_{1}),(q_{2}, a_{2}),…, (q_{t-1}, a_{t-1})]$, where $(q_{i}, a_{i})$ indicates the question and answer in the i-th turn in conversation history, the goal is to verify the rationale for the $q_t$ and give answer $a_t$.

\vspace{-2.2mm}
\subsection{Prompt Design}
\label{ssec:subhead}
\vspace{-1.1mm}

The manual-designed prompt is efficient in many tasks\cite{liu2021gpt}. To enhance the information of input questions and rationales, we propose three prompts considering the connection between conversational questions and rationales, question categories, and contextual information. These prompts are detailed in Table 1.
\vspace{-2.7mm}

\begin{table}[!htbp]
\centering
\caption{Three proposed prompts
}
\vspace{-1mm}

\scalebox{0.90}{

\begin{tabular}{ l | m{5cm}  } 
  \hline
  
  \textbf{Prompt} & \textbf{Text}  \\ 
  \hline

  \colorbox{Question}{Question} \colorbox{Rationale}{Rationale} &  \colorbox{Question}{Question: q} + \colorbox{Rationale}{Rationale: r}  \\ \hline

  \colorbox{Question Description}{Question Description} & \colorbox{Question Description}{Category Question: q + Rationale: r }   \\ \hline

  \colorbox{History}{Conversation History} & \colorbox{History}{Category Question: $q_{t-1}$ +}  \\    
& \colorbox{History}{Rationale: $r_{t-1}$+Answer: $a_{t-1} +$} \\     

 & \colorbox{History}{Category Question: q +} \\&  \colorbox{History}{Rationale: r + Answer:}    \\ \hline
\end{tabular}
}

\vspace{-8.5mm}

\end{table}

\vspace{-0.7mm}

\subsubsection{Question Rationale Prompt}
\vspace{-0.9mm}

The first one is the question rationale relationship prompt. We naturally use a special symbol $[SEP]$ to isolate the question and rationale text, but it lacks prior information and requires fine-tuning to learn the connection. To better activate the pre-trained model's knowledge, we add the words ``Question" and ``Rationale" before the corresponding text to directly establish the connection between the two input texts.

\vspace{-2.6mm}
\subsubsection{Question Description Prompt}
\vspace{-1.3mm}

In the second prompt, we aim to address the impact of question categories on the paraphrasing of the rationale text. The mapping of rationale to the answer often relies on the question category\cite{style}. The rationale may require different types of paraphrasing for various question categories, including but not limited to tense transition, pronoun substitution, and conjunction substitution. For instance, count questions require counting the enumerates in the text and providing the number as the answer. To capture the impact of question categories, we analyze the first word of the questions in the training set and select the most common interrogative words as question categories. We then append the corresponding interrogative word to ``Question" as a question description prompt.


\vspace{-2.6mm}
\subsubsection{Conversation History Prompt}
\vspace{-1.3mm}

 To enhance the fluency and appropriateness of the generated answers, we propose using the conversation history as a prompt for input. The rationale text, extracted from the passage based on the comprehension of the question and conversation history, may be semantically incomplete and lack essential information. By incorporating the conversation history as a prompt, we can provide necessary background information during the pragmatic reasoning process of answer generation.

Additionally, question-answer pairs in the conversation history serve as answer examples to guide the model in generating style-aware answers for different categories of questions. This helps alleviate the burden of the model in learning all question-aware styles\cite{reina}. To implement this, we assemble the corresponding history conversation question-rationale-answer samples with text class labels ``question" and ``answer" and add them to the beginning of the input in history order.



\vspace{-2.6mm}

\subsection{Guided Copy Mechanism}
\label{ssec:subhead}
\vspace{-1.3mm}


We introduce a guided copy mechanism \cite{vinyals2015pointer,pgt,copy_open_qa} that incorporates prompt-enhanced information under self-attention to verify the suitability of extracted rationales. If the rationale is not qualified, the mechanism performs an edit to ensure the answer is fluent and appropriate. By directly referencing the vocabulary from the question and rationale, the copy mechanism tackles issues of incorrect or inconsistent terminology, hence fostering consistent usage across multi-turn dialogues.

\vspace{-2.2mm}

\subsubsection{Multi View Copy Distribution}
\vspace{-0.7mm}

To fully capture the different levels of syntactic and semantic information, we propose that the last decoder should attend to all encoder layers \cite{bert-insight1,bert-insight2}, a concept similar to but distinct from the transparent attention \cite{bapna-etal-2018-training} method, which mainly improves gradient flow. The multi-head of the self-attention mechanism \cite{vaswani2017attention} would learn diverse aspects of the information, and we hope some heads will focus on the prompts. We employ the encoder-decoder attention in the last decoder layer as the copy distribution of the copy mechanism \cite{self_attention_guidance,bi2020palm,wu-etal-2019-transferable}. To allow the model to determine the importance of attention weight from different encoder layers, we assign a learnable weight to it. To denote the copy probability of the token in context, we use:
\vspace{-4.2mm}

\begin{equation}
\alpha_{t, l}=\sum_{i}^N W_{a,i}\operatorname{softmax}\left(\frac{W_s s_t (W_{h,i} h_{l,i})^T}{\sqrt{d_k}}\right)
\end{equation}

\vspace{-1.5mm}
$h_{l,i}$ represents the representation of the $l$-th token in context from the $i$-th encoder, while $s_t$ represents the output representation of the $t$-th token from the decoder. $W_{a,i}$ is a learnable parameter, and $W_s$ and $W_h$ are the query and key parameters, respectively, of the scaled dot-product attention. The final copy distribution for $y_t$ is:
\vspace{-2.5mm}

\begin{equation}
P_{c o p y}(y_t)=\sum_{i: x_i=y_t} \alpha_{t, l}
\end{equation}
\vspace{-5.5mm}
\subsubsection{Verify and Edit}
\vspace{-0.7mm}

The generation coefficient 
$p_{\mathrm{gen}}$ is calculated from the last encoder's input representation $h_l$, with $a_l$ and $s_t$. It controls whether to generate a new word or copy it from the context.

\vspace{-5mm}

\begin{equation}
p_{gen}=\operatorname{sigmoid}\left(w_c\sum_l^Na_lh_l+w_s s_t+b\right)
\end{equation}
\vspace{-2mm}

The final generation probability of $y_t$ is the combination of vocabulary distribution $P_{\text {vocab }}$ and copy distribution $P_{\text {copy}}$:
\vspace{-2mm}

\begin{equation}
P_{\text {vocab }}(y_t)=\operatorname{softmax}\left(W_v s_t+b_t\right)
\end{equation}

\vspace{-5mm}

\begin{equation}
P(y_t)=p_{\text {gen }} P_{\text {vocab }}(y_t)+\left(1-p_{\text {gen }}\right) P_{\text {copy}}(y_t)
\end{equation}


\vspace{-2.7mm}

\section{Experiments}
\label{sec:Experiments}

\vspace{-2mm}

\subsection{Datasets}
\label{ssec:subhead}
\vspace{-1.2mm}

Our experiments are conducted using the CoQA dataset. Specifically, we use the train set for training our model and evaluate it on the dev set. CoQA consists of English dialogues that include questions, answers, and rationale. For our model, we use the question and rationale as the input and the answer as the output label. We consider an answer to be extractive if it overlaps with the rationale, in which case the answer is directly chosen as the rationale. In the train set, which contains 135K samples, 66.8\% of the answers are extractive, meaning they are directly taken from the text, while the remaining 33.2\% require rewriting, which we classify as generative samples. The dev set consists of 5.5k extractive answers and 2.4k generative samples.

\vspace{-2.5mm}

\subsection{Model Setting}
\vspace{-1.2mm}

Our proposed method is evaluated using the T5 base model\cite{t5}, which has 220 million parameters, 6 encoder layers, and 6 decoder layers with 8 heads. We construct our experiments based on the Huggingface Transformers framework\cite{wolf-etal-2020-transformers}. The model is fine-tuned with prompts initialized from the T5 checkpoint using the Adam optimizer with a learning rate of 2e-5 and cross-entropy as the loss function. We train the model for 10 epochs using a batch size of 8. Our experiments were implemented on a V100 featuring 16GB of memory, and an estimated 1.5 hours was required for each epoch of training and evaluation.

\vspace{-2.3mm}

\subsection{Comparing Methods}
\vspace{-1.2mm}

We assess the performance of our T5-based Prompt-Guided Copy Mechanism (PGC-T5) against the following methods:

\begin{itemize}
\vspace{-0.5mm}
  \item Raw: Human annotation of the rationale as the answer
    \vspace{-0.1mm}

  \item Vanilla-T5: Vanilla T5-base model using $[SEP]$ to concatenate the question and rationale. 
    \vspace{-0.1mm}
  \item V-PGNet-T5: T5-based pointer generator, which uses the vanilla encoder-decoder attention as the copy distribution leveraging $[SEP]$ to concatenate the question and rationale.
    \vspace{-0.1mm}
\item P-T5: Vanilla T5-base model with the conversation history prompt. 
\item P-PGNet-T5: T5-based pointer generator, which uses the vanilla encoder-decoder as the copy distribution using the conversation history prompt.
\end{itemize}


  
\vspace{-4mm}

\subsection{Evaluation}
\vspace{-1.2mm}

Our proposed method is evaluated and compared with others using standard CoQA evaluation metrics of EM and F1 scores. Our focus is on the performance of both extractive and generative answers, taking into consideration the dataset distribution. We report three types of EM scores: O-EM for overall samples, G-EM for generative samples, and E-EM for extractive samples. Additionally, we report three types of F1 scores: O-F1 for overall samples, G-F1 for generative samples, and E-F1 for extractive samples.

\vspace{-2.2mm}
\subsection{Main Result}
\label{ssec:subhead}
\vspace{-1.2mm}

Results from our CoQA dataset experiments are captured in Table 2. We observe that only 66.3\% of the human-annotated rationales are considered as answers. While all models perform similarly well in extracting answers, their performance in generative tasks varies. 

The vanilla T5-base model significantly improves the quality of rationales, but at the cost of the extracted answer score. Incorporating the copy mechanism causes a slight drop in the generation score of PGNet-T5. However, by utilizing the design prompt to guide the copy mechanism, PGNet-T5 achieves higher generation scores while preserving extraction scores. Our experiments show that this approach outperforms P-T5, highlighting the effectiveness of the copy mechanism in handling lengthy prompts. Furthermore, PGC-T5, attending to all encoder layers, generates more fluent and appropriate answers. 

\begin{table}[!htbp]
\vspace{-2.7mm}

\caption{The results of compared models on the CoQA dev set. }
\vspace{-1.2mm}
\scalebox{0.76}{
\begin{tabular}{l|cccccc}
\hline \textbf{Models} & \textbf{O-EM} & \textbf{O-F1} & \textbf{G-EM} & \textbf{G-F1} & \textbf{E-EM} & \textbf{E-F1} \\
\hline Raw & $66.3 $ & $73.4$ & $0.0$ & $18.1$ & $\mathbf{95.5}$ & $\mathbf{97.8}$  \\
T5 & $83.6$ & $89.8$ & $59.1$ & $72.9$ & $94.4$ & $97.3$ \\
V-PGNet-T5 & $83.3$ & $89.6$ & $58.3$ & $72.2$ & $94.4$ & $97.3$  \\
P-T5 & $83.9$ & $90.1$ & $60.3$ & $73.7$ & $94.3$ & $97.3$ \\

P-PGNet-T5 & $84.0$ & $90.2$ & $60.8$ & $74.1$ & $94.2$ & $97.3$  \\
PGC-T5 (version3)& $\mathbf{84.2}$ & $\mathbf{90.4}$ & $\mathbf{61.3}$ & $\mathbf{74.8}$& $94.3$ & $97.3$ \\
\hline
\end{tabular}}
\vspace{-1.5mm}

\end{table}
\vspace{-0.9mm}

\begin{table*}[!htbp]
    \caption{Typical edit types in the dev set of CoQA}
    \vspace{-1.2mm}
    \centering
    \scalebox{0.97}{
        \begin{tabularx}{\textwidth}{XXXX}
            \toprule
            \textbf{Type} & \textbf{Question} & \textbf{Extractive Answer} & \textbf{PGC-T5 Answer} \\
            \midrule
            Word Class & How many planets are there away from the Sun? & the fifth planet from the Sun & Five \\
            \hline
            Pronoun & When does he not pay attention to Dan? & when I'm well & when he's well \\
            \hline
            Tense & What did she do first? & taking a nap & took a nap \\
            \hline
            Yes-No & Did she return safely? & Gardner was nowhere to be found & no \\
            \hline
            Simplification & Who are the 2 main actors/actresses in the film? & Ryan Reynolds wonders if marrying his boss, Sandra Bullock & Ryan Reynolds and Sandra Bullock \\
            \hline
        \end{tabularx}%
    }
    \vspace{-4.5mm}
\end{table*}

\vspace{-5.2mm}

\subsection{Ablation Study}
\label{ssec:subhead}
\vspace{-1.5mm}


We evaluate the effectiveness of our proposed prompt through a series of experiments and analyze the results. In Table 4, we conduct a linear upgrade of our prompt and evaluate it using the PGNet-T5 + Multi-view attention. The upgraded versions include the question rationale prompt (Version 1), question description (Version 2), and conversation history prompt (Version 3). Our experiments demonstrate that the model's performance significantly improved with the incorporation of our prompt.


\begin{table}[!htbp]
\caption{The ablation experiments of proposed methods }
\vspace{-1.2mm}
\scalebox{0.85}{
\begin{tabular}{l|cccccc}
\hline \textbf{Models} & \textbf{O-EM} & \textbf{O-F1} & \textbf{G-EM} & \textbf{G-F1} & \textbf{E-EM} & \textbf{E-F1} \\
\hline Base model & $83.2$ & $89.5$ &$57.8$ & $71.8$& $94.4$ & $97.3$ \\
+version 1& $83.7$ & $90.0$ &  $59.3$ & $73.1$ &$94.5$ & $97.4$ \\
+version 2& $83.8$ & $90.0$ &$59.4$ & $73.3$ & $\mathbf{94.5}$ & $\mathbf{97.4}$  \\
+version 3& $\mathbf{84.2}$ & $\mathbf{90.4}$ &$\mathbf{61.3}$ & $\mathbf{74.8}$& $94.3$ & $97.3$  \\
\hline
  \end{tabular}}
  \vspace{-5mm}

\end{table}


\textbf{The question rationale prompt} helps the model identify the input text type and focus on the rationale text.

\textbf{The question description prompt} outperforms the question rationale prompt in generative score while maintaining the same extractive score. The question description prompt encourages the model to generate more fluent and appropriate answers, leading to significant improvements in various question categories as illustrated in Figure 2. Please note that we only show the top 10 categories by quantity in the figure due to space constraints.

\vspace{-2.2mm}

\begin{figure}[!htbp]
\vspace{-2.2mm}

\centering
\includegraphics[width=0.47\textwidth,height=0.30\textwidth]{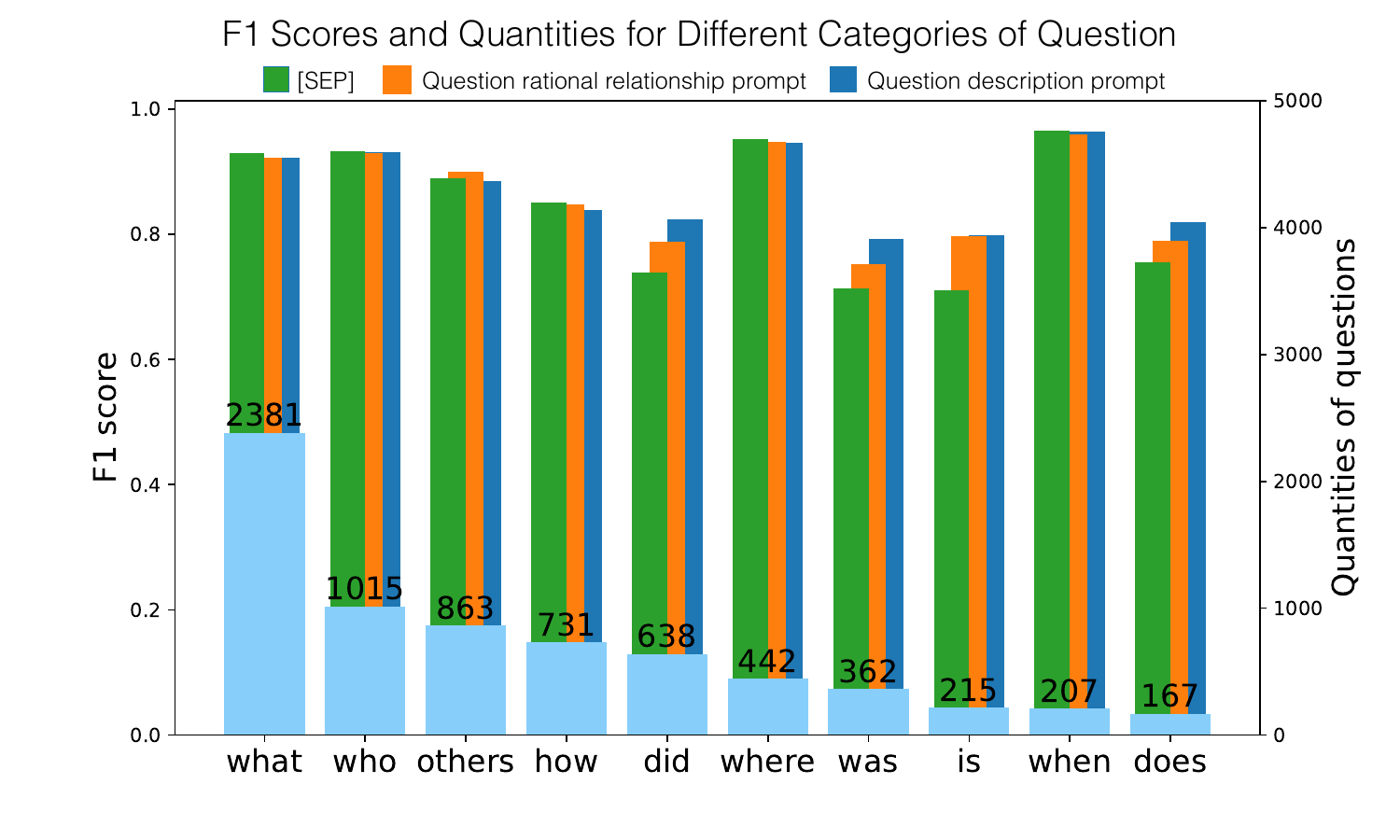}
\vspace{-7.5mm}

\caption{F1 scores and Quantities for question categories. }

\label{fig}
\vspace{-4.5mm}

\end{figure}

{\bf The conversation history prompt} can significantly improve the overall score, with a generative score improvement of 1.5 F1 and 1.9 EM. This prompt can help the model better understand vague questions, handle coreference, and remove redundant answer text in the conversation history. For example, as shown in Table 5, this prompt can drop the redundant text ``overdose of sedatives" from the final answer. Figure 3's attention map provides insights into how the model directs its attention in response to the prompt. The left graph demonstrates the model's ability to distinguish between repeated information and the answer based on the conversation history. The right graph highlights four distinct attentional components of the model, including its focus on the conversation history itself, the connection between the history and the current turn's question, the repetition of answers in response to the current turn's question and history, and the current turn's input.

\begin{table}[!htbp]
\centering

\caption{Example of redundant text removal in CoQA Dev set
}
\vspace{-1.2mm}
\scalebox{0.85}{

\begin{tabular}{ c | m{6cm}  } 
  \hline
  \textbf{Turn} & \textbf{Text}  \\ 
  \hline
  1 & Q: He died of what? \\ & A: an overdose  \\ \hline
  2 & Q: Of what?  \\ & A: sedatives  \\ \hline
  3 & Q: And what else? \\ & R: overdose of sedatives and the surgical anesthetic propofol
  \\ & A: surgical anesthetic propofol
  \\ & A (version 1): overdose of sedatives and the surgical anesthetic propofol \\ &
  A (version 3): surgical anesthetic propofol \\ \hline

\end{tabular}
}

\vspace{-3.7mm}

\end{table}

\begin{figure}[!htbp]
\centering
\includegraphics[width=0.49\textwidth,height=0.245\textwidth]{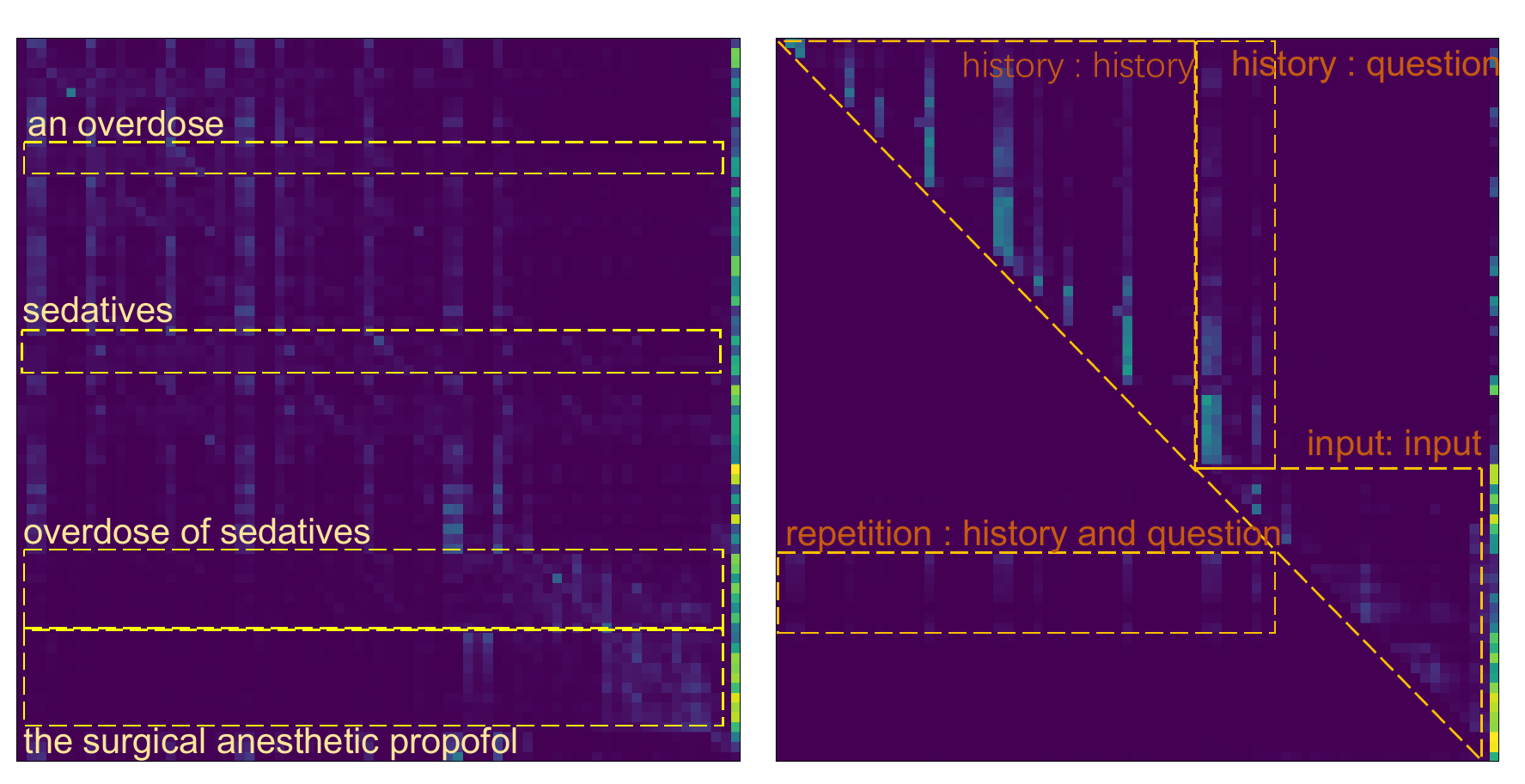}
\vspace{-6mm}

\caption{Visualizing Attention: Two heads of last encoder layer}
\label{fig}
\vspace{-9mm}

\end{figure}

\vspace{-3.5mm}

\subsection{Human Evaluation}
\label{ssec:subhead}
\vspace{-1.2mm}

In our study, we carried out a human evaluation on the predicted development set, assessing the fluency and appropriateness of the generative samples. Table 3 shows five notable improvements in the methods concerning these aspects:

\begin{itemize}
    \item\textbf{Word Class}: The model understands the required word class and refines potential words accordingly.
    \item\textbf{Pronoun}: The model correctly uses pronouns consistent with the question.
    
    \item\textbf{Tense}: The model flexibly changes verb tense in the answer based on the tense of the question.
    \item\textbf{Yes-No}: The model recognizes yes-no questions and provides appropriate answers.
    \item\textbf{Simplification}: The model removes redundant text and improves the answer structure based on the question.
\end{itemize}
    
\vspace{-3.2mm}

\section{Conclusions}
\label{sec:conlusion and future work}
\vspace{-1.7mm}

Our paper proposes a pluggable approach for extractive methods that uses a prompt-guided copy mechanism to improve the quality of extracted answers in Conversational Question Answering. The approach employs prompts to link questions to answers and attention to verify and make necessary edits for the extracted answers to be fluent and appropriate. Our experiments on the CoQA challenge show the effectiveness of our prompt modeling and multi-view attention-based copy mechanism. Further research is needed to verify its generalizability and performance under different scenarios.

\vspace{-3.2mm}

\section{Acknowledgement}
\vspace{-1.7mm}

This paper is supported by the Key Research and Development Program of Guangdong Province under grant No.2021B0101400003. Corresponding author is Jianzong Wang from Ping An Technology (Shenzhen) Co., Ltd (jzwang@188.com).

\bibliographystyle{IEEEtran}
\bibliography{pgc_references} 

\end{document}